\documentclass[journal]{IEEEtran}
\ifCLASSINFOpdf
\else
\fi
\usepackage{amsmath}
\usepackage{amsfonts}
\usepackage{amssymb}

\usepackage[ruled,norelsize]{algorithm2e}

\usepackage{adjustbox}
\usepackage{tikz}
\usepackage{caption}
\usepackage{subcaption}
\usepackage{pgfplots}
\usepgfplotslibrary{groupplots}
\pgfplotsset{width=\linewidth,compat=1.9,
                    every axis/.append style={
                                label style={font=\small},
                                tick label style={font=\small},
                                enlarge x limits={abs=0.40}},
                                /pgfplots/ybar legend/.style={
                                /pgfplots/legend image code/.code={%
                                \draw[##1,/tikz/.cd,yshift=-0.40em]
                                (0cm,0cm) rectangle (3pt,0.8em);},
},
}

\begin{document}
%
\title{DiPSeN: Differentially Private Self-normalizing Neural Networks For Adversarial Robustness in Federated Learning}
%
%
%

\author{Olakunle~Ibitoye,~\IEEEmembership{Member,~IEEE,}
        M.~Omair~Shafiq,~\IEEEmembership{Member,~IEEE,}
        and~Ashraf~Matrawy,~\IEEEmembership{Senior~Member,~IEEE}
\thanks{Olakunle Ibitoye, M.Omair Shafiq, and Ashraf Matrawy are with Carleton School of Information Technology.}

}

\maketitle

\begin{abstract}

The need for robust, secure and private machine learning is an important goal for realizing the full potential of the Internet of Things (IoT). Federated learning has proven to help protect against privacy violations and information leakage. However, it introduces new risk vectors which make machine learning models more difficult to defend against adversarial samples. 

In this study, we examine the role of differential privacy and self-normalization in mitigating the risk of adversarial samples specifically in a federated learning environment. We introduce DiPSeN, a Differentially Private Self-normalizing Neural Network which combines elements of differential privacy noise with self-normalizing techniques. Our empirical results on three publicly available datasets show that DiPSeN successfully improves the adversarial robustness of a deep learning classifier in a federated learning environment based on several evaluation metrics.
\end{abstract}


\ifCLASSOPTIONpeerreview
\begin{center} \bfseries EDICS Category: 3-BBND \end{center}
\fi
%
\IEEEpeerreviewmaketitle

\section{Introduction}
%
%
%
%
\IEEEPARstart{F}{ederated} learning \cite{mcmahan2016federated} is a recent trend in privacy preserving machine learning in which multiple client devices collaboratively participate in a machine learning process without revealing their private data. Each client device maintains a local model, while a master device aggregates the local models from the client devices. Federated learning is a promising solution to the several privacy concerns which limit the acceptability of the Internet of Things (IoT).

While federated learning seeks to solve the problem of privacy violation and information leakage in machine learning (ML), other common challenges such as security and robustness still remain prevalent in federated learning. In this study, we explore the problem of ML robustness in a federated learning setting. ML models are known to be vulnerable to adversarial samples. Adversarial samples are specially crafted data samples, which are designed to alter the training or inference pipelines of a machine learning system in order to impact the reliability of the machine learning system \cite{ibitoye2019threat}.

Various techniques such as adversarial training \cite{kurakin2016adversarial} have been proposed for defending against adversarial samples. Many of these solutions do not address the distributed and autonomous nature of federated learning. Hence, in this study, we propose a technique for defending against adversarial samples, that is suited for a federated learning environment.

Our study combines two techniques - differential privacy and self-normalization - and evaluates their combined effectiveness in improving the adversarial robustness of neural networks in a federated learning environment.

Normalizaton is a technique used in neural networks to dampen the oscillations that occur in the distribution of activations at the output of each neuron or node. When normalization is applied, a neural network model's ability to generalize is significantly improved and the training time is reduced as a result of normalization during the back propagation process. Also, the neural network becomes more resilient to vanishing and exploding gradients as a result of the normalization process. A Self-normalizing Neural Network (SNN) \cite{klambauer2017self} is a type of deep learning model that maintains the stability of the network during the gradient descent process. A self normalizing neural network replaces the standard activation functions in a typical neural network with a specific activation function known as the scaled exponential linear unit (SELU).

Differential Privacy involves the addition of random noise to a data sample to produce anonymity. A differentially private algorithm must provide guarantee that the result of analysis on a certain dataset remains unchanged even if a particular record was absent from the dataset.

We combine the two concepts - differential privacy, and self-normalization - to implement a neural network model that is more robust to adversarial samples. Both concepts respectively address two elements that are known to cause neural network models to be vulnerable to adversarial samples; sensitivity and invariance. Both elements are further discussed in section \ref{archi}.


\textbf{Our Contributions} in this paper are as follows: For our \textbf{ first contribution}, we introduce a neural network model termed "DiPSEN" which demonstrates increased robustness to adversarial robustness in a \textbf{federated learning} setting.

\textbf{For our second contribution} we demonstrate that adversarial robustness can be improved without significantly increasing the computational overhead of the neural network training and inference. We consider this an important criteria for applying federated learning in an IoT environment due to the resource constraint nature of client participants in a federated learning setting.

In our \textbf{third and final contribution}, we demonstrate that our approach is useful in an heterogeneous environment which is typical to IoT environment, by using heterogeneous data samples, including a combination of images and network traffic data. Our results show DiPSEN improves adversarial robustness  with heterogeneous datasets, which are typical for a federated learning environment.

\section{Background}

\subsection{Adversarial Machine Learning}\label{AML}
A machine learning model can be represented as a function  \(f(x)\) that maps an input \(X\) into an output \(y\), where input \(X\) is a vector of numerical values, and output \(y\)   is either a real number \(\) or a label. Since our study here focuses on a classification model, we discuss adversarial examples in the context of a neural network classification model. Given a classifier 
\(f\colon X \mapsto \{1,...,k\} \) where \(X  \in \mathbb{R}^n\). The output of the model is the highest probability from a probability distribution across the labels. The model output could thus be defined as a vector \(y\) which represents the scores such that \(y_k(x) \in [0,1] and \sum_{k=1}^Ky_k(X)=1  \), and such that \(f(x) = argmax_{k\in K y_k} (x) \).

An adversary may induce a wrong prediction from this model by crafting a carefully constructed input \(x'\) which is known as an adversarial sample. The adversarial sample is generated by introducing a small perturbation \(\alpha\) to the input \(x\) such that \(x + \alpha = x'\).

\subsection{Differential Privacy}
The primary goal of differential privacy (DP) in machine learning is to preserve the privacy of the training data by adding some form of noise to induce uncertainty in the data distribution \cite{abadi2016deep}. A typical mechanism for achieving differential privacy as proposed by Dwork et al. \cite{dwork2014algorithmic} approximates a deterministic function \(f:D \rightarrow \mathbb{R}\) which is calibrated to the sensitivity of \(f\). In our study, the framework for generating the differential privacy noise follows two steps. First, the parameters for the additive noise are carefully selected following a Gaussian distribution. A sequential composition of bounded-sensitivity functions is then approximated. More details are provided in section \ref{alg} and in Algorithm \ref{fig:alg1} below. 

\subsection{Self-normalizing Neural Networks}
Neural networks with deep architectures have been known to experience gradient decay, resulting in poor performance. During stochastic gradient descent, the distribution of the weights "W" in the neural network as well as the outputs "x" of each layer are known to vary significantly for every iteration of the stochastic gradient descent process. As a result of the variations, the training process becomes very unstable, hence resulting in saturated activations and consequently introducing the problem of vanishing gradients. Klambauer et al \cite{klambauer2017self} proposed the Self-normalizing Neural Networks (SNN) which is a variant of the ANN that uses a Scaled Exponential Linear Unit (SELU) activation function. The scaled exponential linear unit is shown as:

\[selu(x) = \left\{
                \begin{array}{ll}
                  x & if \, x >0\\ 
                  \alpha^x - \alpha & if \; x <= 0\\
                \end{array}.
              \right.\\
              \]

With the SELU activation function, the mean of the activation output is kept at zero and the variance is kept at one. This allows for deeper neural network architectures to be trained without suffering significant gradient decay. 

\begin{figure*}[!ht]
	\includegraphics[width=0.7\linewidth,keepaspectratio=true]{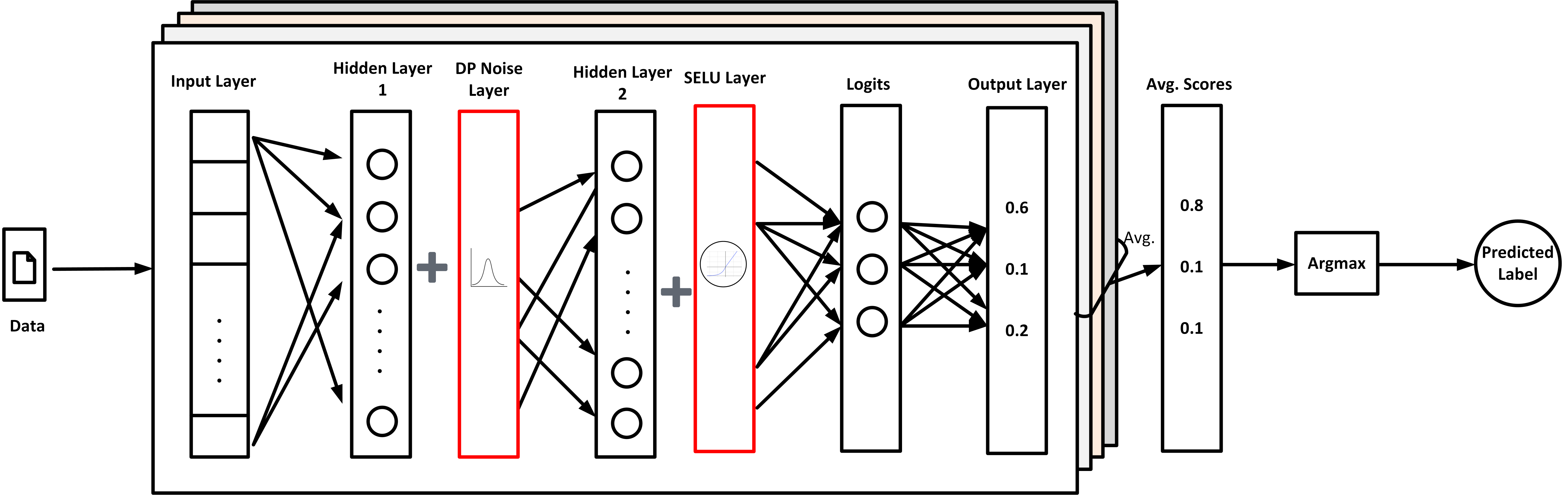}
	\centering
	\caption{DiPSeN Architecture}
	\label{fig:model_archi}
	\centering
\end{figure*}

\subsection{Federated Learning}

In a federated learning setting, we have a number of clients \(C_1,... C_n\) who jointly train a machine learning model \(\theta_g\) while keeping their individual data private. Each of the clients \(C_i\) receives the same initialized copy of the global model \(\theta_g\) from a central server or aggregator \(A\) who initializes the global model \(\theta_g\) and broadcasts it to the client participants. Each client device \(C_1, C_2,... C_n\) stores the model on the device as a local model \(\theta_l\) and then trains their respective local model \(\theta_{l1}, ... \theta_{ln},\) using their own respective private data \(D_1, ... D_n,\).

The federated learning process begins with a round of learning \(t\) in which the aggregator \(A\) broadcasts  \(\theta_g\) to only a few selected client participants. \(C_1,... C_{tn}\). The clients update their respective local models from \(\theta_{li}^t\) to \(\hat{\theta}_{li}^t\) and send it back to the aggregator. The aggregator then updates the global model \(\theta_g\) using a specific aggregation  algorithm such as Federated Averaging (FedAvg) \(\Delta\) \cite{mcmahan2016federated} . 

Multiple rounds of training are conducted until the global model is fully trained. The global model represents an average of the knowledge embodied in all the client devices depending on the aggregation algorithm which was used. In this study, we explore the FedAvg \cite{mcmahan2016federated}, which simply computes the average of the parameters of each of the individual updated local models.

\section{Problem Statement}
Defending against adversarial samples in a federated learning environment is quite challenging due to the distributed and autonomous nature of federated learning. One of the most effective defences against adversarial samples is adversarial training proposed by Madry et al \cite{madry2018towards}.

Adversarial training is usually undertaken as a mimax optimization problem, which updates the weights of the model with regards to the adversarial sample. The approach of adversarial training is to conduct a robust optimization \cite{abou2020investigating}, the effectiveness of which depends on the ability to find the worst case adversarial sample, and make the model robust to this sample. Sinha et al. \cite{sinha2017certifying} have proven that the task of finding  this worst possible adversarial sample is a very difficult optimization task to achieve. 

Adversarial training is impractical in a federated learning setting for some obvious reasons. First, since the data cannot be inspected, it is impossible to set the appropriate bounds for p-norm to perform adversarial training. Secondly, adversarial training requires significant compute compute resources, which is not often feasible on the client participant devices. Thirdly, adversarial training was primarily designed for Independent and Identically distributed (IID) data \cite{jacobsen2019exploiting} which the distributed and autonomous nature of federated learning schemes usually do not provide 
\cite{madry2018towards}.


In our previous work \cite{ibitoye2019analyzing}, we showed that self-normalizing properties of neural networks was a provable means of achieving adversarial robustness. Our previous approach does not apply in this case, since in a federated learning scheme, the training data is not centralized, but distributed across several clients. Hence, we investigate adversarial robustness in a different context.
Compared to our previous work, in this new study, we combine self-normalization with differential privacy noise to improve robustness against adversarial samples.

\section{ Our Proposed Solution - DiPSeN}

In this study, we propose a Differentially Private Self-normalizing Neural Network, which we term as DiPSeN. Our solution works by combining a differential privacy noise layer, with a self-normalizing layer, to improve adversarial robustness.

\subsection{Methodology} \label{archi}
Our model \(f(x)\) has a scoring function denoted as \(Q(x)\) which maps the input \(x\) to a probability distribution of labels \(y\). The adversarial vulnerability of the model \(f(x)\) is based on the unbounded sensitivity of \(Q(x)\)  to changes in the inputted \(x\) measured according to the p-norm change \cite{lecuyer2019certified}. The excessive invariance of the model further contributes to the adversarial vunerability of the model \cite{jacobsen2018excessive}.  Our goal is to reduce both the sensitivity of scoring function \(Q(x)\), and the invariance of the model \(f(x)\). To reduce the sensitivity of the scoring function, we add a differentially private noise layer drawn from a Gaussian distribution as shown in Fig. \ref{fig:model_archi}. To reduce the invariance in the model parameters, we stabilize the model by adding a self-normalizing layer to the architecture.

\subsection{Differentially Private Noise Layer}

The connection between differential privacy (DP) and adversarial robustness was studied by Lecuyer et al. \cite{lecuyer2019certified}, who demonstrated that a classifier which satisfies DP with respective to atomic units such as pixels, can be leveraged for adversarial robustness.

 For our study, we incorporate a noise layer that satisfies differential privacy requirements. The noise layer generates a noise input with zero mean, using the Gaussian distribution. A Gaussian noise distribution is preferably selected since it inherently has lower variance, compared to other distribution such as laplace distribution, based on our objective of further reducing the invariance in the subsequent layers. As shown in Fig. \ref{fig:model_archi}, the noise layer is placed directly after the first hidden layer. Similar to the work of \cite{lecuyer2019certified}, the noise distribution is directly proportional to the sensitivity of the scoring function \(Q(x)\) from the output of the first hidden layer. The result is a randomized scoring function \(\hat{Q}(x)\) with an output that is less sensitive to changes in the input, unlike the initial scoring function which was highly sensitive.  To further reinforce the model to the vulnerability effects of invariance, we apply a selu layer to the subsequent layers, which we discuss in the next subsection \ref{selu}.


\subsection{SELU layer} \label{selu}

Our goal in adding the SELU layer is to mitigate the excessive invariance which is known to cause adversarial vulnerability in our neural network model. The intuition behind self-normalization is to keep the mean and the variance as close to 0 and 1 respectively throughout each layer of the neural network. The SELU layer comprises of a Scaled Exponential Linear Unit (SELU) activation function followed by an AlphaDropout layer \cite{klambauer2017self} which randomly sets the neurons to a specified value instead of a zero value which is typical for the standard dropout, enabling us to retain the mean and variance at 0 and 1 respectively. For initializing the weights, a Lecun Uniform Initializer \cite{lecun2012efficient} is utilized.

\section{Threat Model}

\begin{figure*}[h]
	\includegraphics[width=0.7\linewidth,keepaspectratio=true]{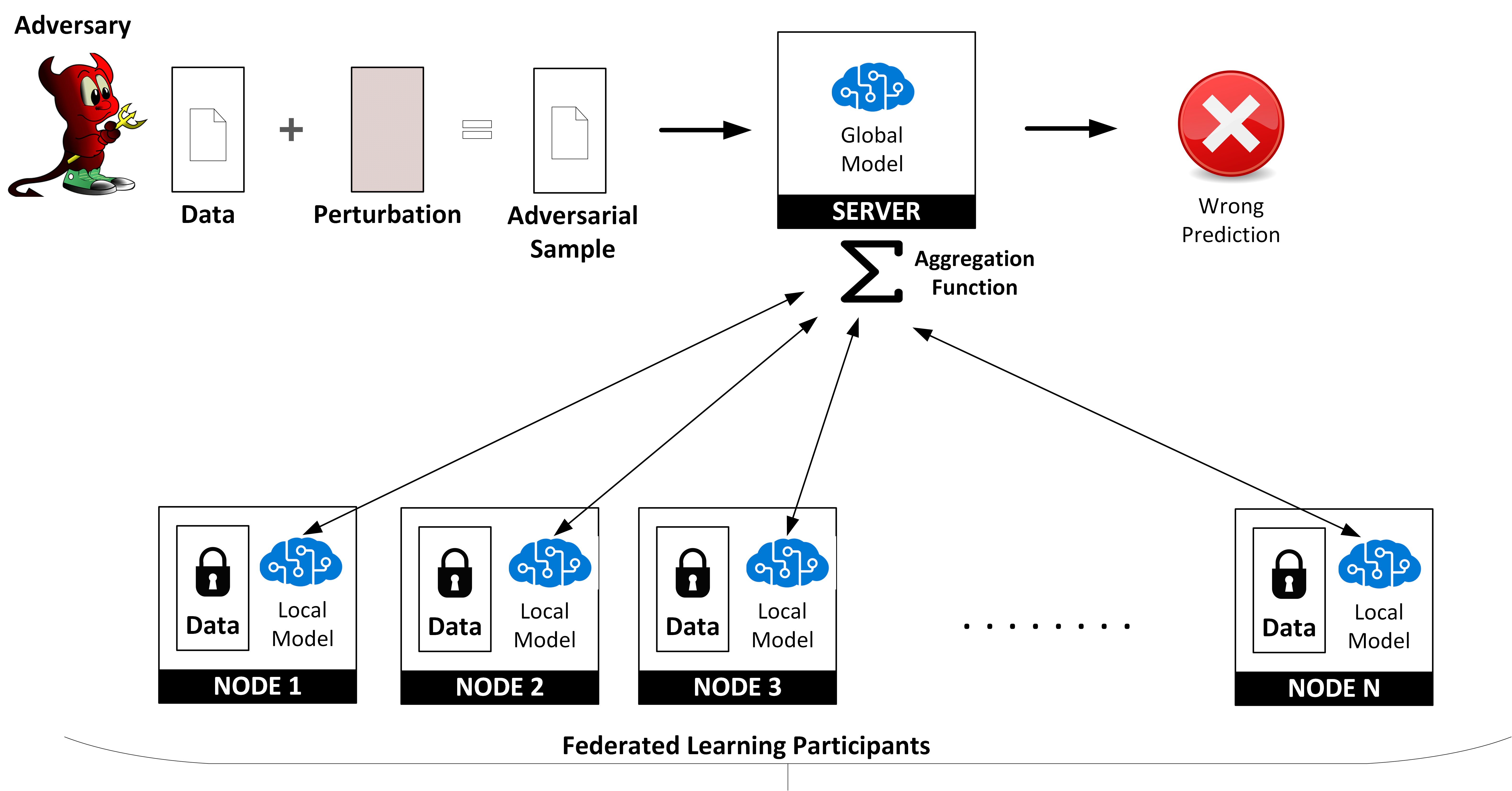}
	\centering
	\caption{Threat Model}
	\label{fig:threat_model}
	\centering
\end{figure*}

 Fig. \ref{fig:threat_model} below depicts our threat model for this study. Our study considers an adversary who alters the inference data to craft adversarial samples with the intent of misleading the machine learning model into making a wrong prediction. This is termed an evasion attack which occurs during the prediction phase of machine learning and differs from a poisoning attack which occurs during the training phase in which the adversary seeks to poison the training data. 
 
Our study assumes an untargeted attack. Untargeted attacks, typically aim to reduce the overall accuracy of the global model without any specific misclassification outcome. This differs from a targeted attack in which the adversary's aim is to achieve a specific misclassification outcome on selected examples, even though the global model maintains a good accuracy for the majority of other samples. 
 
For our study, we assume a white box attack, with the adversary being able to inspect and observe the model parameters during the entire attack process. We consider that a federated learning scheme could have three possible scenarios in which the adversary could have knowledge about the model parameters. Firstly, the adversary may have a complete knowledge of the current deep learning model, and can observe or inspect the model parameters during the training and prediction phases. This is known as a white box attack. A second possibility is an orange-box attack, in which the adversary only has access to a stale version of the deep learning model. The third possibility is a black-box attack, meaning that the adversary is unable to observe the model parameters at any stage and has no knowledge about the model.

\section{Experimental Setup}

\subsection{Details, overall architecture of the experimental setup } The experiments were carried out on a local deep learning workstation with processor details - Intel(R) core(TM) i7-9700 CPU, 8 cores, @3.00GHz. The PC has a disc storage of 500 Gb SSD and 32GHz DDR4 Random Access Memory (RAM). For the software, the code was written in Python 3.7 using the Pycharm integrated development Environment and with the Anaconda data science distribution. The deep learning model was implemented using Tensorflow V2 machine learning framework with the inbuilt keras API.  All experiments were run locally using a separate command line console for each experiment. All tools, frameworks and datasets utilized are open source or publicly available.

The adversarial samples and the federated learning parties were generated using the IBM adversarial Robustness Toolbox (ART) \cite{nicolae2018adversarial} and federated learning \cite{ludwig2020ibm} frameworks respectively. 

\subsection{Neural Network Model and Federated learning Scheme}
We build a Convolutional Neural Network (CNN) model, intialized it at the aggregator and broadcasted to the client participants. The CNN model consists of one input layer, two hidden layers and one fully connected (Dense) layer. Each hidden layers consist of a convolution layer and a pooling layer for extracting hidden representations from the dataset. The noise layer is applied after the first hidden layer while the Self-normalizing layer is added after the second hidden layer as shown in Fig. \ref{fig:model_archi}.    

The federated learning setting is shown in Fig. \ref{fig:threat_model}. For the experiment, we implemented a federated learning scheme with one aggregator, and 200 client participants or nodes. The aggregator itself does not participate in the training process but initiates the parameters of the model and sends a copy of the model to a randomly selected number of clients to participate in the first round. The client participants update their local copy of the model and send the updated version to the aggregator who aggregates the parameters of each clients model. The aggregator then sends the updated model to another selected number of clients to participate in the second round. The process continues until an accurate model is obtained.

\subsection{Dataset, feature engineering and data pre-processing}
We use three  datasets for our study. One image classification dataset, one fingerprint biometric dataset and one network traffic dataset. For the first image dataset, the MNIST dataset \cite{lecun2010mnist} is used which contains 60,0000 training examples and 10,000 test examples. Each instance of the dataset represents an image which consists of 28 X 28 pixels and is loaded as a  matrix with values ranging from 0-255. After loading the dataset, the images are scaled to values between 0 and 1 by dividing the matrix by 255. All included features for the MNIST dataset are used in building the neural network model. 

The second image dataset is the Sokoto Conventry Fingerprint dataset (SocoFing) \cite{shehu2018detection} which is a biometric fingerprint. The dataset consists of 6000 fingerprint images from 600 African datasets and contains unique attributes of each participant in the study. The labels for the dataset include gender, hand and finger name. It also includes synthetically altered versions with three different levels of alteration for obliteration, central rotation, and z-cut. Each of the sample in the dataset is a (28x28) px grey-scale pixel image of one of the ten digits 0 - 10. Hence the input vector into our neural network model has a shape (28x28x28) pixels. 

The third dataset is the CTU-13 dataset \cite{garcia2014empirical} which is a network traffic dataset of botnet traffic consisting of real botnet traffic combined with background traffic and normal traffic. The original CTU-13 dataset contains 16 features out of which we use 6 features which are considered most relevant to the study. The remaining features were omitted since they had less impact on the output and were likely to result in overfitting of the model. The dataset was standardized by scaling them to values between 0 and 1. To prepare the CTU-13 dataset for the convolution layer, the dataset was reshaped into 3 dimensions to represent a shape of (28,28,1).

\subsection{Algorithmic details, and complexity}\label{alg}

The pseudocode for our solution is presented in Algorithm \ref{fig:alg1} which illustrates the steps taken to build the model and initiate the federated learning across the multiple clients and aggregate the various local models at the server end. The parameters for the neural network model are grouped into a single input \(\theta\) with respect to the loss function \(L\). The Gaussian noise distribution is represented as \(N (0,S^{2}_{f}.\sigma^2) )\) with a mean of 0 and a standard deviation of \(S^{2}_{f}\sigma\). The SELU layer is shown to map the mean and variance of the activation in each layer with the mapping that maintains \((\mu,v)\) and  \((\overline{\mu},\overline{v})\) as close to 0 as possible. The complexity of the algorithm is simplified by restricting it to a real-valued function.

\makeatletter
\newcommand{\removelatexerror}{\let\@latex@error\@gobble}
\makeatother

\begin{figure}
 \removelatexerror
  \begin{algorithm}[H]
  \caption{DiPSEN Algorithm Details}
  	\label{fig:alg1}
  Initialize \(\theta_t\) at Server\\
  Initialize {number of hidden layers L, weights w}\;
  Add  input layer, activation layer, dropout layer\\
  
  \For {i in L - 1}
  {
\textbf{Add DP Noise Layer}\;
\(\textsl{g}_t \leftarrow 1/L (\sum \textsl{g}_t(x_i) + N (0,S^{2}_{f}.\sigma^2)\)\\
\textbf{Add SELU Layer}\;
\(g(\Omega) = {(\mu,v)| \mu \in [\mu_{\min},\mu_{\max}],v \in [v_{\min},v_{\max}]}\)\\
}
 Add output layer, softmax activation layer\\

  \For{each training round t = 1,2,3...}
  {Select \(m = C * K\) clients, \(C \in (0,1)\) clients\\
  Download \(\theta_t\) to each client \(k\)\\
  \For{each client \(k \in m\)}{
  Wait for Client to synchronize\\
  Aggregate local model}
  }

  \end{algorithm}
\end{figure}

\subsection{Generating The Adversarial Samples}\label{generate_adv_samples}
We generate our adversarial samples using the Adversarial Robustness Toolbox (ART) \cite{nicolae2018adversarial} framework which is provided by IBM and is made available for public use.

The first method we use in generating the adversarial examples for the IoT dataset is the Fast Gradient Sign Method (FGSM). This method performs a one step gradient update along the direction of the sign of gradient for every input in the dataset. \cite{Goodfellow2015ExplainingAH}. The second method is the Basic Iteration method (BIM) which runs a finer optimization of the FGSM with minimal smaller changes for multiple iterations \cite{kurakin2016adversarial}. In each iteration, the each feature of the input values is clipped to avoid too large a change on each feature. The third method is the Projected Gradient Descent (PGD) which is also a variation of the FGSM attack but omits the random start feature of the FGSM \cite{madry2018towards}. All three methods are model dependent methods and rely on the model gradient.

In  our experiments, we craft the adversarial examples with the intent of misleading the classifier without any specific target labels being specified. The epsilon number represents the maximum perturbation for the adversarial attack and in our experiment, we select an epsilon of 0.3.

\section{Experimental Approach}

In this section we evaluate the effectiveness of a differentially private self-normalizing neural network on the adversarial robustness of deep learning models. Our evaluation criteria are carefully selected with regards to crucial requirements in a federated learning setting. Firstly, we bear in mind that federated learning datasets are typically non -IID and often times heterogeneous data samples. Our evaluation dataset thus consists of three types of datasets: a baseline MNIST image classification, fingerprint biometric detection dataset, and a network intrusion detection dataset. Since the client participants in a federated learning environment are often resource constraint, we evaluate out solution based on the power consumption attributes. Finally, we evaluate our solution based on the impact of the noise on the prediction accuracy. We answer the following three questions in our evaluation steps.

\begin{itemize}
    \item What is the impact of the additional layers on the model accuracy?

  \item How effective is our solution for resource constrained devices?
  
    \item Does our solution scale across different domains such as image classification and network security?   
\end{itemize}

\subsection{Approach}
In carrying out this study, we utilize a four-step approach. First, we implement a deep learning classifier in a federated learning scheme using a Convolution Neural Network (CNN) model. We evaluate the accuracy of the CNN model using three datasets. The resultant accuracy from the CNN model is what we term the 'Basic CNN Accuracy".

Next, we enhance the CNN model by adding a DP noise layer, and a SELU layer. This enhanced CNN model is our proposed solution which we refer to as "DiPSEN". We evaluate the classification accuracy of DiPSEN with adversarial free samples. The resultant accuracy from this improved model with adversarial free samples is what we term the "Baseline Accuracy". The intent of this step is to verify that the two additional layers do not negatively impact on the accuracy of the model. 

In the third step, we measure the computational overhead of DiPSEN. This is an important metric since the client devices in a federated learning setting are assumed to be regular day to day devices such as IoT devices,  with limited computing power. The intent of this step is to determine if the two additional layers - the dp noise layer and the SELU layer - contribute to an excessive computational overhead, which would make our solution impracticable in an IoT environment. 

In the fourth step, we generate adversarial samples and then evaluate the adversarial robustness of DIPSEN. We compare the results with the basic CNN model. We evaluate the adversarial robustness across three heterogeneous datasets to verify the applicability of our solution with heterogeneous data samples.

\subsection{Evaluation Metrics}
We use five classification metrics in our evaluation to determine the accuracy of the classifier. These are accuracy (acc), precision (pre), recall (re), F-1 score (f1) and support (sup). The computational overhead is evaluated based on the average time per training step.

\section{Results}

\subsection{Basic CNN Accuracy}\label{basic_cnn_acc}
In the first subsection, we conduct an evaluation of a deep learning classifier using a basic CNN model with all three datasets. We establish that the CNN model accurately predicts the various class labels for each dataset. Our result in table \ref{basic_cnn_acc} shows that the CNN model achieves an accuracy of 97\%, 82\% and 92\% on the MNIST, SocoFing and CTU-13 datasets respectively. This indicates that the CNN model performs well on a variety of datasets without any adversarial samples. 

\begin{table}[ht]
\caption{Basic CNN Model Accuracy}
\label{tab_basic_cnn_acc}
\begin{center}
\begin{tabular}{|c|c|c|c|c|c|c|}
\hline
S/N & Dataset & acc & pre & rec & f1 & sup \\
\hline
\hline
1 & MNIST(Image) & 97\% & 96\% & 96\% & 97\% & 97\%\\
\hline
2 & SocoFing(Biometrics) & 82\% & 82\% & 82\% & 83\% & 82\%\\
\hline
3 & CTU-13(Botnet) & 92\% & 93\% & 93\% & 93\% & 93\%\\
\hline

\end{tabular}
\end{center}
\end{table}

\subsection{DIPSEN Baseline Accuracy} 
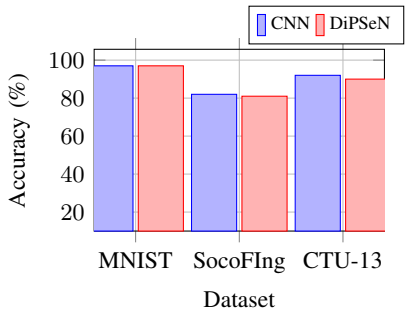
\begin{figure}[h!]
	\centering
	\pgfplotsset{width=3cm,height=4cm,compat=1.15}
	\begin{tikzpicture}[baseline]
	\begin{axis}[
	ybar,
	ymin=10,
	grid=major,
	ylabel={Accuracy (\%)},
	xlabel={Dataset},
	xticklabels={MNIST,SocoFIng,CTU-13},
	legend style={font=\scriptsize,at={(0.8,1.24)},
		anchor=north,legend columns=-1},        
	xtick=data,
	bar width=6mm,
	width=0.3\textwidth,
	]
	\addplot table[x=K,y=Accuracy]{\modelAccuracy};
	\addplot table[x=L,y=Accuracy2]{\modelAccuracyNoise};
	\legend{CNN, DiPSeN}
	\end{axis}
	\end{tikzpicture}
	\caption{DiPSeN Baseline Accuracy}
	\label{fig:noise_acc}
\end{figure}

In the previous experiment, we established a functional CNN model, with high accuracy across all three datasets. In this second step, we implemented the additional layers to the basic CNN model namely the DP noise layer and the SELU layer, which were aimed at providing adversarial robustness. We term this new model as "DiPSEN" which is our proposed solution for improving adversarial robustness. We evaluated the impact of the added layers that made up DiPSEN and compared with the basic CNN model in terms of accuracy. The results shown in Fig. \ref{fig:noise_acc} indicate that the additional layers for adversarial robustness do not adversely impact the classification accuracy of DiPSEN. We find that the classification accuracy of our DiPSEN solution is comparable to that of the basic CNN model.

\pgfplotstableread[row sep=\\,col sep=&]{
	K    & Accuracy\\
	1   & 97 \\
	2   & 82\\
	3  & 92 \\
}\modelAccuracy

\pgfplotstableread[row sep=\\,col sep=&]{
	L    & Accuracy2\\
	1   & 97 \\
	2   & 81\\
	3  & 90 \\
}\modelAccuracyNoise

\subsection{Computational Overhead of DiPSEN vs basic CNN model.}

In the third step, we compare the computational overhead of DiPSEN with the computational overhead of a basic CNN model. We find that DiPSEN does not significantly add to the computational overhead required for training the neural network model. Our evaluation of the basic CNN computational overhead using our Intel(TM) Core i7 processor with GPU RTX 2070 takes an average of 9.6s, 11.5s and 4.8s per training step for the MNIST, SocoFing and CTU-13 datasets respectively. With DiPSEN, a slight increase of 0.9\% increased overhead on average with 98\% confidence interval is observed across all three datasets.  As such, DiPSEN does not significantly degrade performance of the neural network training. 

\pgfplotstableread[row sep=\\,col sep=&]{
	K    & time\\
	1   & 9.6 \\
	2   & 11.5\\
	3  & 4.8 \\
}\computationOverhead

\pgfplotstableread[row sep=\\,col sep=&]{
	K    & time\\
	1   & 9.7 \\
	2   & 11.6\\
	3  & 4.9 \\
}\computationOverheadNoise

\begin{figure}[h!]
	\centering
	\pgfplotsset{width=3cm,height=4cm,compat=1.15}
	\begin{tikzpicture}[baseline]
	\begin{axis}[
	ybar,
	ymin=0,
	grid=major,
	ylabel={Computation Overhead (sec)},
	xlabel={Dataset},
	xticklabels={MNIST,SocoFIng,CTU-13},
	legend style={font=\scriptsize,at={(0.8,1.24)},
		anchor=north,legend columns=-1},        
	xtick=data,
	bar width=6mm,
	width=0.3\textwidth,
	]
	\addplot table[x=K,y=time]{\computationOverhead};
	\addplot table[x=K,y=time]{\computationOverheadNoise};
	\legend{CNN, DiPSeN}
	\end{axis}
	\end{tikzpicture}
	\caption{Computational Overhead}
	\label{fig:overhead}
\end{figure}
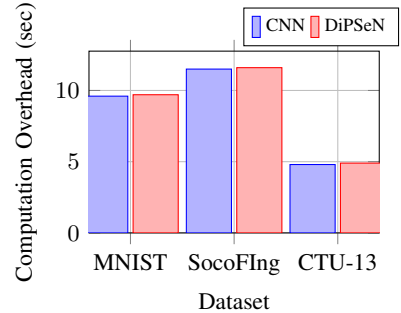

\subsection{Adversarial Robustness of DiPSEN compared to basic CNN model.}
\begin{figure*}
	\begin{center}
		\pgfplotsset{width=3cm,height=4cm,compat=1.15}
		\begin{tikzpicture}
		
		\begin{groupplot}[
		group style={
			group name=my plots,
			group size=3 by 1,
			horizontal sep=1.7cm,
			vertical sep=2.3cm,
		},
	   ybar,
	   xtick={1,2,3},
		xticklabels={FGSM,BIM,PGD}, 
		ylabel = {Accuracy (\%)},
		xlabel={Attach Method},
		grid=major,
		ymax=100,
		ymin=0,
		width=0.3\linewidth
		]
		
		\nextgroupplot[legend to name={CommonLegend},legend style={legend columns=2}]
		\addplot table[x=K,y=time]{\mnistAccuracyBaseline};
		\addplot table[x=K,y=time]{\mnistAccuracyDiPSeN};
		\addlegendentry{CNN}
		\addlegendentry{DiPSeN}
		\nextgroupplot
		\addplot table[x=K,y=time]{\mnistAccuracyBaseline};
		\addplot table[x=K,y=time]{\mnistAccuracyDiPSeN};
		\nextgroupplot
		\addplot table[x=K,y=time]{\mnistAccuracyBaseline};
		\addplot table[x=K,y=time]{\mnistAccuracyDiPSeN};
		\end{groupplot}
		\path (my plots c1r1.south east) -- node[yshift=-18mm]{\ref{CommonLegend}} (my plots c3r1.south west);
		\node[text width=6cm,align=center,anchor=north] at ([yshift=-8mm]my plots c1r1.south) {\subcaption{MNIST Dataset}\label{fig:mnist_robustness}};
		\node[text width=6cm,align=center,anchor=north] at ([yshift=-8mm]my plots c2r1.south) {\subcaption{SocoFing Dataset\label{fig:SocoFing_robustness}}};
		\node[text width=6cm,align=center,anchor=north] at ([yshift=-8mm]my plots c3r1.south) {\subcaption{CTU-13 Dataset\label{fig:CTU-13_robustness}}};
		\end{tikzpicture} 
	\end{center}
	\caption{Adversarial robustness of CNN vs. DiPSEN.}\label{fig:my_label}
\end{figure*}
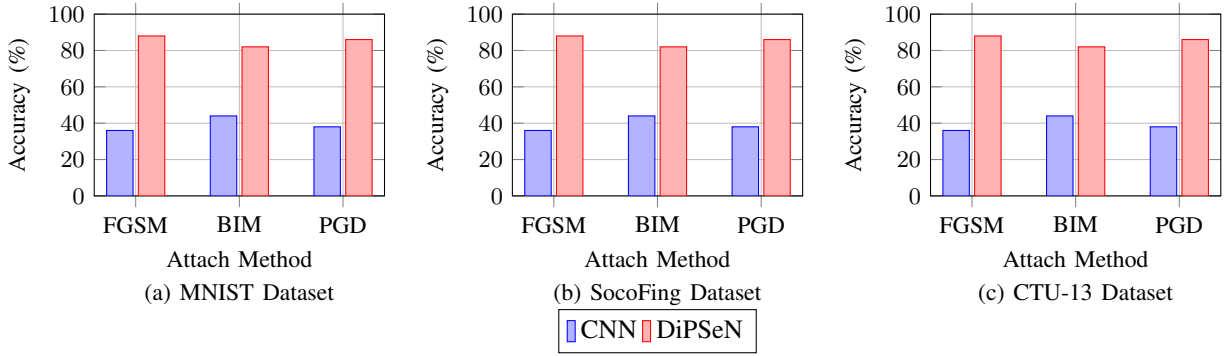

\begin{figure*}[h]
    \centering
    \includegraphics[width=\linewidth,keepaspectratio=true]{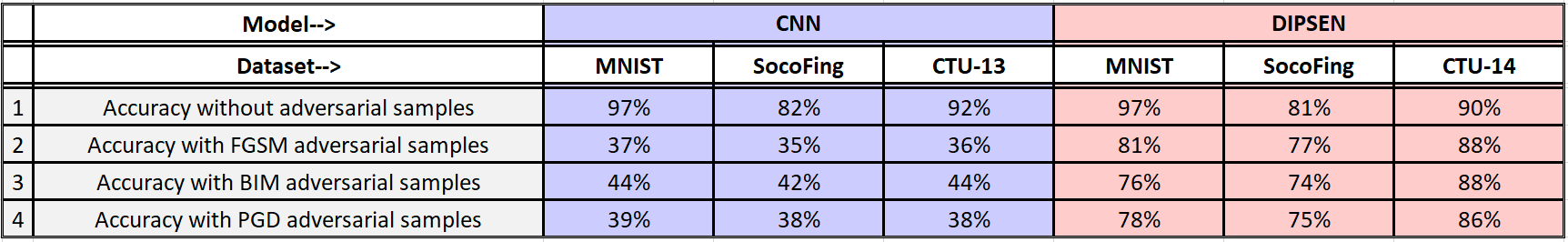}
    \caption{Summary table of results}
    \label{fig:summary_table}
\end{figure*}

In the fourth and final step, We generated adversarial samples using the three adversarial attack methods described in section \ref{generate_adv_samples} namely FGSM, BIM and PGD. we then compared the adversarial robustness of our proposed solution - DiPSEN, with the adversarial robustness of the basic CNN model by. The adversarial robustness is evaluated by testing the prediction accuracy of the models using the generated adversarial samples. The results shown in Fig. \ref{fig:mnist_robustness}, \ref{fig:SocoFing_robustness} and \ref{fig:CTU-13_robustness} show that DiPSEN has an improved adversarial robustness compared to the basic CNN model. A summary table of all results is shown in Fig. \ref{fig:summary_table}.

\pgfplotstableread[row sep=\\,col sep=&]{
	K    & time\\
	1   & 37 \\
	2   & 44\\
	3  & 39 \\
}\mnistAccuracyBaseline

\pgfplotstableread[row sep=\\,col sep=&]{
	K    & time\\
	1   & 81 \\
	2   & 76\\
	3  & 78 \\
}\mnistAccuracyDiPSeN

\pgfplotstableread[row sep=\\,col sep=&]{
	K    & time\\
	1   & 35 \\
	2   & 42\\
	3  & 38 \\
}\mnistAccuracyBaseline

\pgfplotstableread[row sep=\\,col sep=&]{
	K    & time\\
	1   & 77 \\
	2   & 74\\
	3  & 75 \\
}\mnistAccuracyDiPSeN

\pgfplotstableread[row sep=\\,col sep=&]{
	K    & time\\
	1   & 36 \\
	2   & 44\\
	3  & 38 \\
}\mnistAccuracyBaseline

\pgfplotstableread[row sep=\\,col sep=&]{
	K    & time\\
	1   & 88 \\
	2   & 82\\
	3  & 86 \\
}\mnistAccuracyDiPSeN

\section{Related Work}
Some solutions proposed Byzantine-resilient defenses as one approach to defend against adversarial samples using mean estimation \cite{yin2018byzantine}. The study performed by Fang et al. \cite{fang2020local} demonstrated that this approach was ineffective against model poisoning attacks with the context of federated learning. Other approaches include the use of data shuffling \cite{chen2018draco}. This approach was also tested for model poisoning attacks, and does not apply to federated learning since the data is distributed and private for each participating client.

Another approach for creating robustness in neural network models, involves the randomization of the model \cite{xie2017mitigating}. By randomizing the model, some element of noise is introduced to the model which creates a smoothing effect on the prediction function. This has been shown to empirically result in a more robust model. Other best effort defenses include model distillation \cite{papernot2016effectiveness} and gradient masking \cite{gao2017deepcloak}.  Most of these defenses have been proven to be insufficient in various ways, and none of them has been proven to defend against adversarial samples in federated learning. The most similar to our work is the study by \cite{lecuyer2019certified} which adopts elements of differential privacy to achieve differential adversarial robustness. 

Data sanitization methods \cite{cretu2008casting} have been proposed specifically for data poisoning attacks, with the aim of removing suspected poisoned or anomalous data from the training set. Other defenses have improved on data sanitization to incorporate advanced techniques such as robust statistics \cite{shen2019learning}. Since data sanitization requires access to the client data which is not feasible in a federated learning setting, this defense technique is not suitable. A similar defense technique to data sanitization is network pruning \cite{wang2019neural} which removes activation units that have not been active on clean data. This method also requires data access to dataset which is representative of the global dataset.



From our literature review, we discovered that no researcher has evaluated the impact of differential privacy noise with self-normalization on the robustness of neural networks in a federated learning scheme. Our study goes further by expanding the context to heterogeneous dataset samples within the context of IoT network security.  Hence our study is novel and offers a useful contribution in understanding the security of machine learning and artificial intelligence in IoT network security.

\section{ Discussion}

\textbf{What defines an adversarial robust model?}
The challenge of finding an adversarial sample is an optimization problem in which the adversary attempts to find a very small change to the input \(x\) that will result in a change to the prediction label. This optimization problem is further challenged by the fact that there is a constraint regarding the amount of change that the adversary can make to input \(x\). The amount of allowed change to the input \(x\) is measured by the p-norm of the change and is represented as \(\|\alpha\|_p\ = (\sum_{i=1}^n |\alpha|^p)^{1/p}\). For a given p-norm \(\|\alpha\|_p\), a data sample \(X  \in \mathbb{R}^n\),  and a given model \(f(x)\)an adversary, an adversary is said to have successfully crafted an adversarial sample if they can compute \(\alpha\) such that \(f(x) + \alpha \neq f(x)\). An adversarial sample is usually a plausible input, such as a data sample which appears valid, while resulting in a wrong prediction output. An adversarially robust model must then be a model whose output is insensitive to small changes in the input \(x\). 

\textbf{How does differential privacy noise improve adversarial robustness?}
The relationship between differential privacy noise and adversarial robustness has been explored in previous works \cite{lecuyer2019certified}. Our approach leverages on this with additional steps to account for the limitations in a federated learning setting.  Adversarial samples rely on perturbations which are generated and added to the original data sample. These perturbations are carefully computed using constrained optimization methods such as fast gradient sign method \cite{papernot2016limitations} in a manner that maximizes the loss function subject to a norm constraint. Our approach begins primarily with transforming the scoring function of our neural network model into a randomized scoring function. This is intended to reduce the sensitivity of the scoring function to perturbed inputs, or adversarial samples. 


\textbf{How does self-normalization improve adversarial robustness?}
We complement the differential private noise layer with a self-normalizing layer. The self-normalizing layer incorporates a scaled exponential linear unit (SELU) activation function which has been shown to elicit three distinct properties. First, the SELU activation function sets to control the average learning rate (u) of the network using negative and positive values. Secondly, The SELU maintains a fixed point in the neural network with the aid of a continuous curve. Thirdly, the saturation region dampens the variance alpha (learning rate)  and the positive slope augments the alpha (learning rate). These three properties help to reduce the invariance of the neural network model \cite{klambauer2017self}.

\textbf{Limitations of our study}
Our solution combines the randomization effects of differential privacy noise with the stabilization effects of the SELU layer to produce a more adversarial robust neural network model. As a potential limitation in our study, the prediction becomes randomized, thereby introducing a different problem of intepretability to the neural network model. This will be explored in a future study.

\section{Conclusion}
In a typical Internet of Things (IoT) environment, federated learning helps to protect against information leakage and privacy violations. However, the distributed and autonomous nature of federated learning makes it difficult to defend against adversarial samples.

We introduced  a solution termed "DiPSEN" which combines a differential privacy noise layer with a self-normalizing layer. The addition of a noise layer randomizes the scoring function, making it less sensitive to perturbed inputs. The self-normalizing (SELU) layer helps reduce the invariance of the model. We demonstrate that both factors help improve the adversarial robustness of the model. We further demonstrate that our solution is effective across heterogeneous datasets, and computationally effective for resource constrained devices, thereby making it suitable for application in a typical IoT environment.


%

\appendices


\section*{Acknowledgment}

This work was supported by the Natural Sciences and Engineering Research Council of Canada (NSERC) through the NSERC Discovery Grant program.

\ifCLASSOPTIONcaptionsoff
  \newpage
\fi

\bibliographystyle{IEEEtran}
\bibliography{main}

\end{document}